\documentclass[runningheads]{llncs}
\usepackage[T1]{fontenc}
\usepackage{graphicx}
\usepackage{orcidlink}
\usepackage{amsmath}
\usepackage{amsfonts}
\usepackage{subcaption}
\usepackage{placeins}
\usepackage{float}
\usepackage{tikz}
\usepackage{hyperref}
\usepackage{algorithm}
\usepackage{enumitem}
\usepackage{comment}
\usepackage{booktabs}
\usepackage[noend]{algpseudocode}
\usetikzlibrary{arrows.meta, positioning}

\begin{document}

\title{A Comparative Study of Adversarial Robustness in CNN and CNN-ANFIS Architectures}
\titlerunning{Adversarial DCNFIS}

\author{
Kaaustaaub Shankar\inst{1}\orcidlink{0009-0000-8970-3746}
 \and
Bharadwaj Dogga\inst{1}\orcidlink{0009-0002-5915-1367} \and
Kelly Cohen\inst{1}\orcidlink{0000-0002-8655-1465}
}

\institute{
College of Engineering and Applied Science, University of Cincinnati, Cincinnati, OH 45219, USA\\
\email{shankaks@mail.uc.edu, doggabj@mail.uc.edu, cohenky@ucmail.uc.edu}
}

\authorrunning{K.Shankar et al.}

\maketitle

\begin{abstract}
Convolutional Neural Networks (CNNs) achieve strong image classification performance but lack interpretability and are vulnerable to adversarial attacks. Neuro-fuzzy hybrids such as DCNFIS replace fully connected CNN classifiers with Adaptive Neuro-Fuzzy Inference Systems (ANFIS) to improve interpretability, yet their robustness remains underexplored. This work compares standard CNNs (ConvNet, VGG, ResNet18) with their ANFIS-augmented counterparts on MNIST, Fashion-MNIST, CIFAR-10, and CIFAR-100 under gradient-based (PGD) and gradient-free (Square) attacks. Results show that ANFIS integration does not consistently improve clean accuracy and has architecture-dependent effects on robustness: ResNet18-ANFIS exhibits improved adversarial robustness, while VGG-ANFIS often underperforms its baseline. These findings suggest that neuro-fuzzy augmentation can enhance robustness in specific architectures but is not universally beneficial.
\end{abstract}

\FloatBarrier
\section{Introduction}
CNNs have become the dominant paradigm for image classification, yet their internal representations and decision processes remain largely opaque, raising concerns for safety-critical domains that require accountability and trust. This black-box behavior has motivated growing interest in explainable and trustworthy AI, particularly in settings such as medical imaging, cybersecurity, and malware detection \cite{yesmin2025fuzzy}, where understanding why a model makes a prediction is as important as the prediction itself. Hybrid approaches that combine deep learning with fuzzy logic have emerged as a promising direction, using fuzzy rules and membership functions to provide human-interpretable decision boundaries while preserving powerful feature extraction \cite{brosolo2025road}.

Within this context, neuro-fuzzy architectures such as the Deep Convolutional Neuro-Fuzzy Inference System (DCNFIS) replace conventional fully connected layers in CNNs with an Adaptive Neuro-Fuzzy Inference System (ANFIS), yielding linguistic rule-based classifiers on top of deep feature maps. Recent work shows that fuzzy logic and neuro-fuzzy models can improve interpretability without severely degrading predictive performance, and can be successfully deployed in real-world regression and classification tasks \cite{Ujong2025Developmen}\cite{yabana2025comparative}. These results position CNN-ANFIS hybrids as attractive candidates for applications that demand both accuracy and transparent decision-making.

At the same time, adversarial attacks such as Fast Gradient Sign Method (FGSM) and Projected Gradient Descent (PGD) have revealed that standard CNNs are highly vulnerable to imperceptible perturbations \cite{waghela2024robust}, prompting extensive research on defensive strategies and adversarial training schemes \cite{villegas2024evaluating}\cite{dai2024adversarial}. Studies on robust training, including hybrid FGSM-PGD procedures and architectural innovations such as sparse mixture-of-experts layers \cite{pavlitska2025robust}, highlight that robustness depends strongly on both training regime and model design. Yet the adversarial robustness of interpretable hybrids like DCNFIS remains poorly understood; this work therefore presents a comparative study of CNN and CNN-ANFIS architectures under gradient-based and gradient-free attacks in white-box and black-box settings, aiming to clarify how interpretability-oriented fuzzy components interact with adversarial robustness.

\FloatBarrier
\section{Related Work}
\subsection{Fuzzy Logic and its Derived Systems}
Fuzzy set theory was introduced by Lotfi Zadeh \cite{ZADEH1965338} as a mathematical concept that allows elements to have partial membership in multiple sets, more closely resembling human reasoning under uncertainty. This can be expressed as:  
\begin{equation}
\mu_{\tilde{A}}(x) \in [0, 1], \quad \forall x \in X
\end{equation}  
where $\mu_{\tilde{A}}(x)$ is the membership function of element $x$ in fuzzy set $\tilde{A}$.
Using this new way of thinking, systems capable of universally approximating nonlinear functions \cite{Castro1995} have been created. Two widely used fuzzy inference systems (FIS) are Mamdani \cite{MAMDANI19751} and Takagi-Sugeno-Kang (TSK) \cite{TSK} and these systems use linguistic terms (low, medium, high) to define both rules and membership functions. 

However, FISs suffer from the curse of dimensionality where an increase in inputs results in an exponential increase in the size of the model. This limits the use of pure fuzzy systems to low-dimensional problems, making them unsuitable for tasks such as image classification. To handle this, an Adaptive Neuro Fuzzy Inference System (ANFIS) \cite{ANFIS} was created which integrates neural network learning algorithms with the structure of a fuzzy inference system allowing for the system to handle larger dimension problems

\subsection{Convolutional Neural Networks and Neuro-Fuzzy Hybrids}

Convolutional Neural Networks (CNNs) are a class of deep learning models specifically designed for processing grid-like data, such as images, through the principle of locality. The key idea behind CNNs is the use of convolutional layers, where small learnable filters, called kernels, slide over the input to extract local patterns such as edges, textures, or shapes. These features are then progressively combined through multiple convolutional and pooling layers to form hierarchical representations, capturing increasingly complex patterns. After feature extraction, fully connected layers perform classification based on the learned representations. Over the years, CNN architectures have evolved from simple models like LeNet \cite{LeNet} for digit recognition, to deeper and more sophisticated networks such as AlexNet \cite{alexnet}, VGG \cite{vgg}, ResNet \cite{resnet}, and DenseNet \cite{densenet}, which incorporate innovations like deeper layers, skip connections, and dense connectivity to improve performance and training stability. CNNs have become the dominant approach in computer vision due to their ability to automatically learn relevant features and generalize across complex image datasets. However, these models are inherently uninterpretable, as the learned kernel features are abstract and the classification aspect is not human-understandable.

To address this limitation, neuro-fuzzy hybrid approaches have been proposed. One such approach replaces the dense layers used for classification in CNNs with an ANFIS, resulting in models like DCNFIS that provide greater interpretability in both feature extraction and classification \cite{dcnfis}. Despite their promise, the robustness and safety of these hybrid models under adversarial attacks remain largely unexplored, providing a strong incentive to investigate their vulnerabilities.

\subsection{Adversarial Attacks}
Since models, whether neural networks or fuzzy-based systems, learn to recognize patterns from the training data, these patterns can be exploited by carefully crafted perturbations to cause incorrect predictions. Such manipulations, known as adversarial attacks, subtly modify inputs in a way that is often imperceptible to humans but can dramatically affect model outputs. These attacks can be categorized into black-box attacks, white-box attacks, and gray-box attacks. For this paper, we focus on both black-box and white-box attacks. Black-box attacks assume that the attacker has no knowledge of the model’s structure or parameters, whereas white-box attacks assume full access to both \cite{adversarialsurvey}. These attacks can also be categorized based on whether they rely on gradient information. Gradient-based attacks use the model’s gradients to craft perturbations that maximize the loss and are typically performed in a white-box setting. In contrast, gradient-free attacks do not require access to gradients and instead rely on model outputs, queries, or random search strategies to generate adversarial examples, making them suitable for black-box scenarios.

Gradient-based attacks, also called white-box attacks, leverage knowledge of the model's internal parameters and gradients to identify input perturbations that maximize prediction errors. These attacks can be highly effective because they directly exploit the model’s learned representations. On the other hand, non-gradient-based attacks, often referred to as black-box or score-based attacks, do not require access to gradients. Instead, they rely on model outputs, query strategies, or random search to identify adversarial inputs, making them particularly relevant for models with partially non-differentiable components, such as neuro-fuzzy hybrids.

Among gradient-based attacks, Projected Gradient Descent (PGD) \cite{pgd} has become a widely used benchmark for evaluating model robustness, due to its iterative, first-order optimization of the input to maximize loss. In the category of black-box, score-based attacks \cite{squareattack}, the Square Attack has emerged as an efficient and effective method that relies solely on model outputs without access to gradients.

\subsubsection{PGD}
PGD \cite{pgd} is an iterative, gradient-based adversarial attack that aims to find perturbations $\delta$ which maximize the model's loss while staying within a specified bound $\epsilon$. The general procedure can be summarized as the follows:

\begin{algorithm}[htbp]
\caption{Projected Gradient Descent (PGD)}
\begin{algorithmic}[1]
\Require Input $x$, true label $y$, model $f$, loss function $\mathcal{L}$  
\Require Step size $\alpha$ (the amount of perturbation added at each iteration)  
\Require Perturbation bound $\epsilon$ (maximum allowed perturbation magnitude)  
\Require Number of iterations $T$  

\State Initialize perturbation: $\delta_0 \gets 0$ (or random within $\epsilon$-ball)  

\For{$t = 0$ to $T-1$}
    \State Compute gradient: $g_t \gets \nabla_x \mathcal{L}(f(x + \delta_t), y)$
    \State Update perturbation: $\delta_{t+1} \gets \delta_t + \alpha \cdot \text{sign}(g_t)$
    \State Project perturbation to $\epsilon$-ball: $\delta_{t+1} \gets \text{clip}(\delta_{t+1}, -\epsilon, \epsilon)$
\EndFor

\State Generate adversarial example: $x_{\text{adv}} = x + \delta_T$
\end{algorithmic}
\end{algorithm}
\subsubsection{Square Attack}
Square Attack \cite{squareattack} as a black-box gradient free adversarial attack generates perturbations by modifying randomly selected square regions of the input to increase the model's loss.
\begin{algorithm}[H]
\caption{Square Attack}
\begin{algorithmic}[1]
\Require Input $x$, label $y$, model $f$, perturbation bound $\epsilon$, max queries $Q$
\State Initialize $x_{\text{adv}} \gets x$
\For{$q = 1$ to $Q$}
    \State Randomly choose a square region in $x_{\text{adv}}$
    \State Apply a random perturbation within $[-\epsilon, \epsilon]$
    \State Clip result to the $\epsilon$-ball around $x$
    \If{loss increases}
        \State Accept the perturbation
    \EndIf
\EndFor
\State \Return $x_{\text{adv}}$
\end{algorithmic}
\end{algorithm}

\FloatBarrier
\section{Methodology}
To make our experiment procedure more robust, we will be training a standard ConvNet, a VGG-like model, and Resnet18 along with their ANFIS variants on 4 different datasets: MNIST, FashionMNIST \cite{fashion}, CIFAR-10 \cite{krizhevsky2009learning}, and CIFAR-100 \cite{krizhevsky2009learning}. 

\begin{table}[htbp]
\centering
\renewcommand{\arraystretch}{1.25}
\setlength{\tabcolsep}{8pt}

\begin{tabular}{l c c}
\toprule
\textbf{Backbone} & \textbf{Feature Dim} & \textbf{Classifier} \\
\midrule
ConvNet    & $n_{\text{feat}}(H,W)$ & FC / ANFIS \\
VGG        & $256$                  & FC / ANFIS \\
ResNet18   & $512$                  & FC / ANFIS \\
\bottomrule
\end{tabular}
\caption{ANFIS replaces FC in CNN}
\end{table}

The ANFIS architecture used in this paper is identical to the one described in Yeganejou et al \cite{dcnfis} and $n_{\text{rules}} = 20$ was used for all variants and all datasets.

We then apply PGD and Square Attack on varying epsilon levels and then calculate the accuracy on the attacked data. This is done through the TorchAttacks library \cite{kim2020torchattacks}.

For the PGD and Square attacks, we used the epsilon values shown in Table~\ref{table:epsilon_hyperparameters}. The alpha value ($\alpha$) for PGD was then derived using the epsilon value as $\frac{1}{35} \cdot \epsilon$ to scale the attacks better while the number of queries for square attack was kept at $n=1000$.

\begin{table}[htbp]
\centering
\renewcommand{\arraystretch}{1.2}
\setlength{\tabcolsep}{10pt}

\begin{tabular}{|c|c|}
\hline
\textbf{Dataset} & \textbf{Epsilon Values} \\
\hline
MNIST & $[0.1,\ 0.2,\ 0.3]$ \\
\hline
Fashion-MNIST & $[0.1,\ 0.2,\ 0.3]$ \\
\hline
CIFAR-10 & $[\frac{2}{255}, \ \frac{8}{255}, \ \frac{16}{255}]$ \\
\hline
CIFAR-100 & $[\frac{2}{255}, \ \frac{8}{255}, \ \frac{16}{255}]$ \\
\hline
\end{tabular}

\caption{Epsilon values used for each dataset.}
\label{table:epsilon_hyperparameters}
\end{table}
To evaluate each of the models fairly given that the complexities of their datasets differ, we will be using a robustness ratio defined as such:

$$\text{Robustness Ratio} = \frac{\text{Adversarial Accuracy}}{\text{Clean Accuracy}}$$

\newpage
\section{Results}
\begin{table}[htbp]
    \centering
    \begin{tabular*}{\textwidth}{l @{\extracolsep{\fill}} c c c c @{\extracolsep{0pt}}}
        \toprule
        \textbf{Model} & \textbf{MNIST} & \textbf{Fashion-MNIST} & \textbf{CIFAR10} & \textbf{CIFAR100} \\
        \midrule
        ConvNet & 99.42\% & 92.49\% & 79.37\% & 45.78\% \\
        VGG & 99.39\% & 92.95\% & 85.63\% & 49.09\% \\
        ResNet18 & 99.41\% & 91.39\% & 82.52\% & 52.98\% \\
        ConvNet-ANFIS & 99.36\% & 92.22\% & 77.41\% & 47.06\% \\
        VGG-ANFIS & 99.51\% & 93.23\% & 85.73\% & 54.74\% \\
        ResNet18-ANFIS & 99.49\% & 91.62\% & 80.69\% & 49.95\% \\
        \bottomrule
    \end{tabular*}
    \caption{Basic Test Accuracy Comparison across Datasets}
    \label{tab:accuracy_comparison}
\end{table}

\begin{figure}[htbp]
    \centering
    \includegraphics[width=0.9\textwidth]{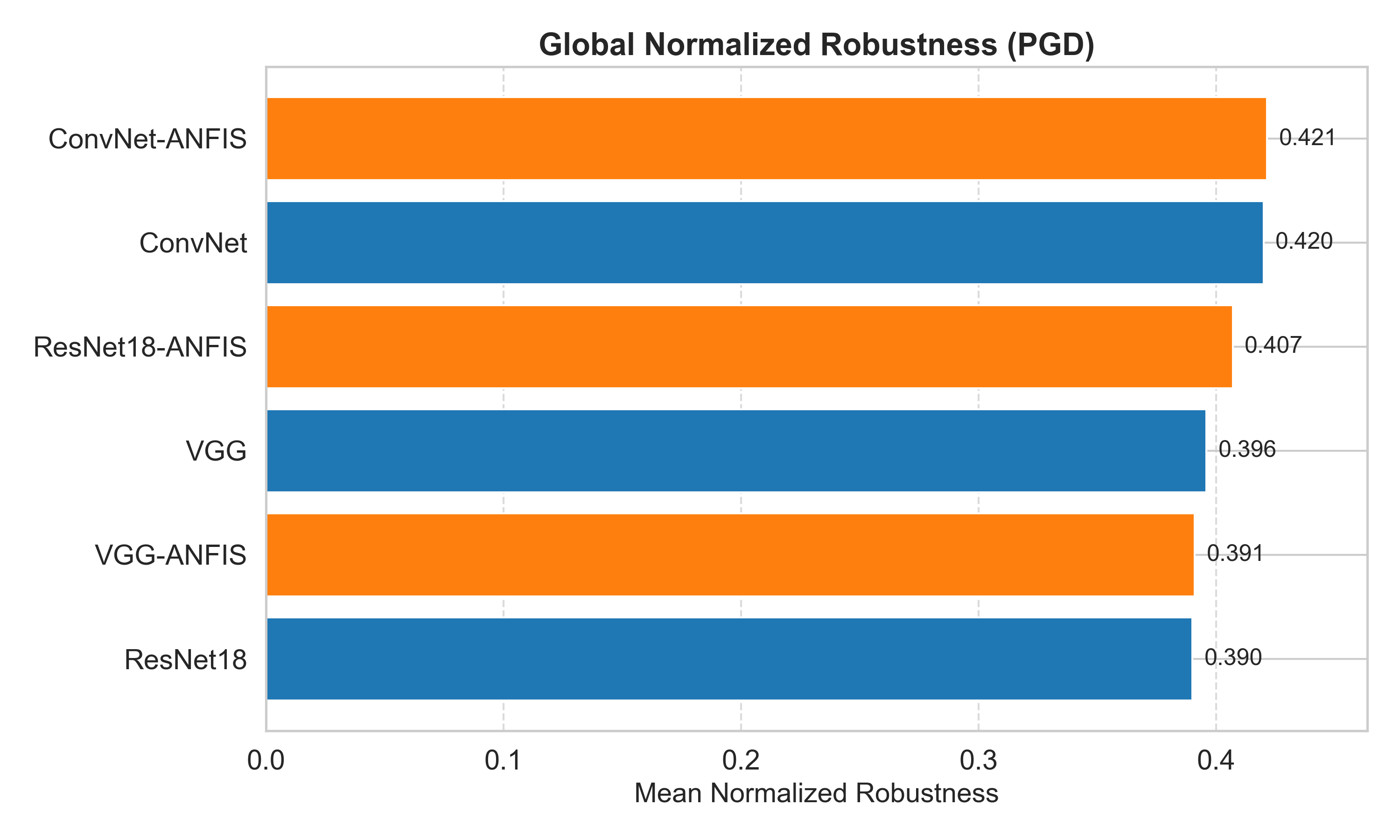}
    \caption{Normalized Robustness for PGD Attacks}
    \label{fig:normalized_robustness_pgd}
\end{figure}
\begin{table}[htbp]
    \centering
    \begin{minipage}{0.48\textwidth}
        \centering
        \caption*{MNIST}
        \begin{tabular}{l c c c}
            \toprule
            \textbf{Model} & $\mathbf{0.1000}$ & $\mathbf{0.2000}$ & $\mathbf{0.3000}$ \\
            \midrule
            ConvNet & 98.85\% & 98.42\% & 97.67\% \\
            VGG & 98.99\% & 98.68\% & 98.20\% \\
            ResNet18 & 91.36\% & 87.97\% & 84.14\% \\
            ConvNet-ANFIS & 98.59\% & 97.95\% & 96.94\% \\
            VGG-ANFIS & 99.07\% & 98.56\% & 97.93\% \\
            ResNet18-ANFIS & 93.33\% & 90.84\% & 87.10\% \\
            \bottomrule
        \end{tabular}
    \end{minipage}%
    \hfill%
    \begin{minipage}{0.48\textwidth}
        \centering
        \caption*{Fashion-MNIST}
        \begin{tabular}{l c c c}
            \toprule
            \textbf{Model} & $\mathbf{0.1000}$ & $\mathbf{0.2000}$ & $\mathbf{0.3000}$ \\
            \midrule
            ConvNet & 34.38\% & 21.36\% & 14.22\% \\
            VGG & 25.87\% & 17.58\% & 12.04\% \\
            ResNet18 & 25.43\% & 15.73\% & 10.68\% \\
            ConvNet-ANFIS & 40.56\% & 25.85\% & 16.68\% \\
            VGG-ANFIS & 25.40\% & 16.04\% & 10.79\% \\
            ResNet18-ANFIS & 30.16\% & 19.40\% & 12.43\% \\
            \bottomrule
        \end{tabular}
    \end{minipage}
    \caption{PGD Attack Accuracy Comparison for MNIST and Fashion-MNIST}
    \label{tab:mnist_fashion_pgd}
\end{table}

\begin{table}[htbp]
    \centering
    \begin{minipage}{0.48\textwidth}
        \centering
        \caption*{CIFAR-10}
        \begin{tabular}{l c c c}
            \toprule
            \textbf{Model} & $\mathbf{0.0078}$ & $\mathbf{0.0314}$ & $\mathbf{0.0627}$ \\
            \midrule
            ConvNet & 32.94\% & 20.60\% & 10.94\% \\
            VGG & 37.56\% & 19.95\% & 8.26\% \\
            ResNet18 & 39.68\% & 26.60\% & 13.65\% \\
            ConvNet-ANFIS & 29.55\% & 19.35\% & 10.73\% \\
            VGG-ANFIS & 36.14\% & 19.85\% & 8.48\% \\
            ResNet18-ANFIS & 39.65\% & 26.66\% & 14.64\% \\
            \bottomrule
        \end{tabular}
    \end{minipage}%
    \hfill%
    \begin{minipage}{0.48\textwidth}
        \centering
        \caption*{CIFAR-100}
        \begin{tabular}{l c c c}
            \toprule
            \textbf{Model} & $\mathbf{0.0078}$ & $\mathbf{0.0314}$ & $\mathbf{0.0627}$ \\
            \midrule
            ConvNet & 12.60\% & 6.90\% & 3.61\% \\
            VGG & 12.53\% & 5.25\% & 2.32\% \\
            ResNet18 & 14.60\% & 7.89\% & 3.68\% \\
            ConvNet-ANFIS & 11.45\% & 6.11\% & 2.81\% \\
            VGG-ANFIS & 14.08\% & 6.07\% & 2.32\% \\
            ResNet18-ANFIS & 13.54\% & 6.99\% & 3.33\% \\
            \bottomrule
        \end{tabular}
    \end{minipage}
    \caption{PGD Attack Accuracy Comparison for CIFAR-10 and CIFAR-100}
    \label{tab:cifar_pgd}
\end{table}

\begin{figure}[htbp]
    \centering
    \includegraphics[width=0.9\textwidth]{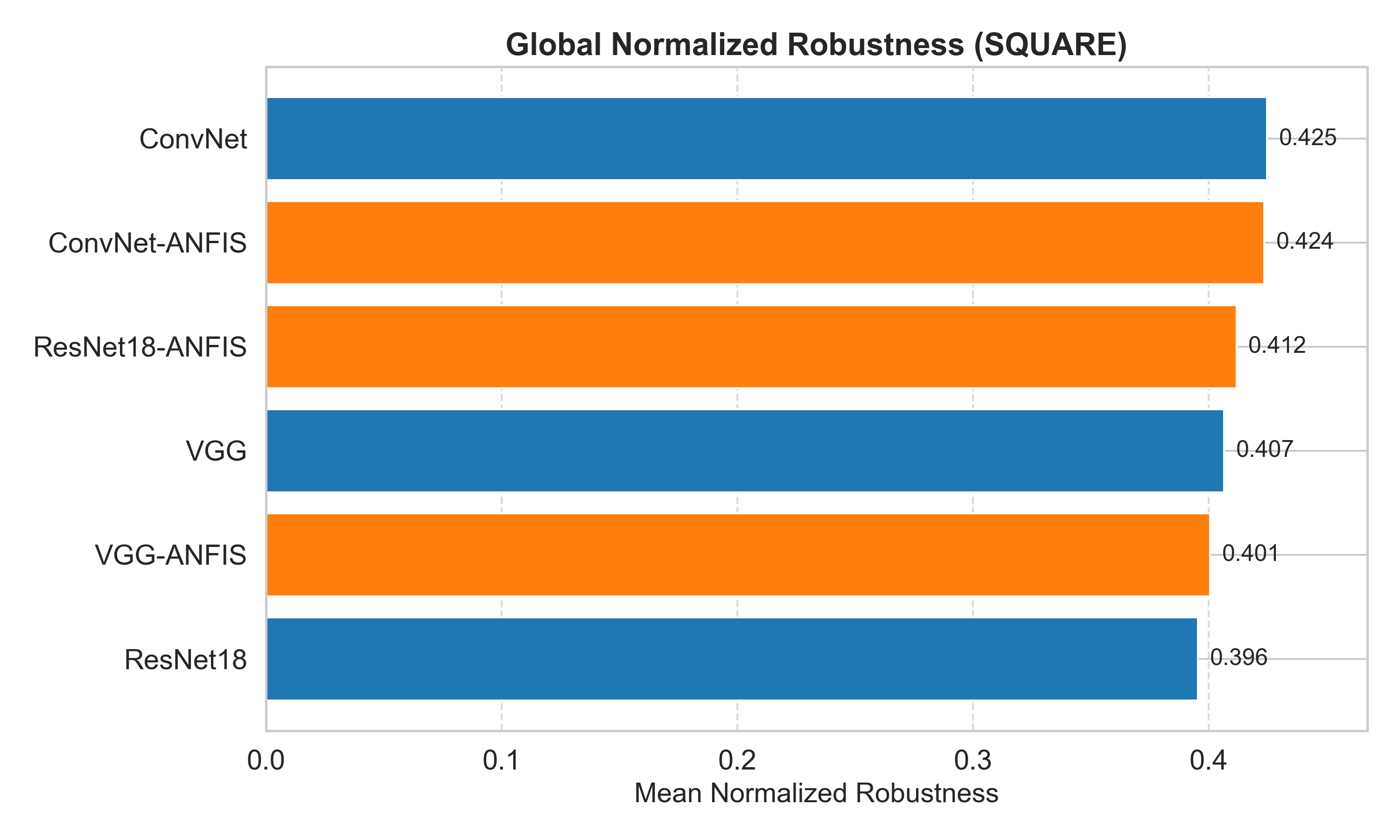}
    \caption{Normalized Robustness for Square Attacks}
    \label{fig:normalized_robustness_square}
\end{figure}

\begin{table}[htbp]
    \centering
    \begin{minipage}{0.48\textwidth}
        \centering
        \caption*{MNIST} 
        \begin{tabular}{l c c c}
            \toprule
            \textbf{Model} & $\mathbf{0.1000}$ & $\mathbf{0.2000}$ & $\mathbf{0.3000}$ \\
            \midrule
            ConvNet & 98.81\% & 98.17\% & 97.38\% \\
            VGG & 98.91\% & 98.57\% & 97.95\% \\
            ResNet18 & 90.78\% & 87.10\% & 82.90\% \\
            ConvNet-ANFIS & 98.39\% & 97.65\% & 96.53\% \\
            VGG-ANFIS & 98.92\% & 98.40\% & 97.73\% \\
            ResNet18-ANFIS & 92.84\% & 90.13\% & 85.99\% \\
            \bottomrule
        \end{tabular}
    \end{minipage}%
    \hfill%
    \begin{minipage}{0.48\textwidth}
        \centering
        \caption*{Fashion-MNIST} 
        \begin{tabular}{l c c c}
            \toprule
            \textbf{Model} & $\mathbf{0.1000}$ & $\mathbf{0.2000}$ & $\mathbf{0.3000}$ \\
            \midrule
            ConvNet & 34.40\% & 22.13\% & 15.50\% \\
            VGG & 25.47\% & 17.67\% & 12.52\% \\
            ResNet18 & 25.61\% & 16.29\% & 11.75\% \\
            ConvNet-ANFIS & 40.18\% & 26.16\% & 17.55\% \\
            VGG-ANFIS & 24.97\% & 16.08\% & 10.85\% \\
            ResNet18-ANFIS & 30.22\% & 20.09\% & 13.43\% \\
            \bottomrule
        \end{tabular}
    \end{minipage}
        \caption{Square Accuracy Comparison for MNIST and Fashion-MNIST}
            \label{tab:mnist_fashion_side_by_side_square}

\end{table}

\begin{table}[htbp]
    \centering
    \begin{minipage}{0.48\textwidth}
        \centering
        \caption*{CIFAR-10} 
        \begin{tabular}{l c c c}
            \toprule
            \textbf{Model} & $\mathbf{0.0078}$ & $\mathbf{0.0314}$ & $\mathbf{0.0627}$ \\
            \midrule
            ConvNet & 31.16\% & 23.51\% & 15.28\% \\
            VGG & 36.92\% & 24.79\% & 13.63\% \\
            ResNet18 & 37.68\% & 28.79\% & 18.65\% \\
            ConvNet-ANFIS & 27.24\% & 21.30\% & 13.93\% \\
            VGG-ANFIS & 35.43\% & 24.11\% & 13.23\% \\
            ResNet18-ANFIS & 37.34\% & 28.93\% & 19.10\% \\
            \bottomrule
        \end{tabular}
    \end{minipage}%
    \hfill%
    \begin{minipage}{0.48\textwidth}
        \centering
        \caption*{CIFAR-100} 
        \begin{tabular}{l c c c}
            \toprule
            \textbf{Model} & $\mathbf{0.0078}$ & $\mathbf{0.0314}$ & $\mathbf{0.0627}$ \\
            \midrule
            ConvNet & 10.66\% & 7.08\% & 4.10\% \\
            VGG & 11.43\% & 6.35\% & 3.26\% \\
            ResNet18 & 13.35\% & 8.59\% & 4.81\% \\
            ConvNet-ANFIS & 9.70\% & 6.47\% & 3.71\% \\
            VGG-ANFIS & 13.37\% & 7.28\% & 3.30\% \\
            ResNet18-ANFIS & 12.11\% & 7.75\% & 4.39\% \\
            \bottomrule
        \end{tabular}      
    \end{minipage}
        \caption{Square Accuracy Comparison for CIFAR-10 and CIFAR-100.}    \label{tab:cifar_side_by_side_square}
\end{table}

\begin{figure}[htbp]
    \centering
    \includegraphics[width=0.9\textwidth]{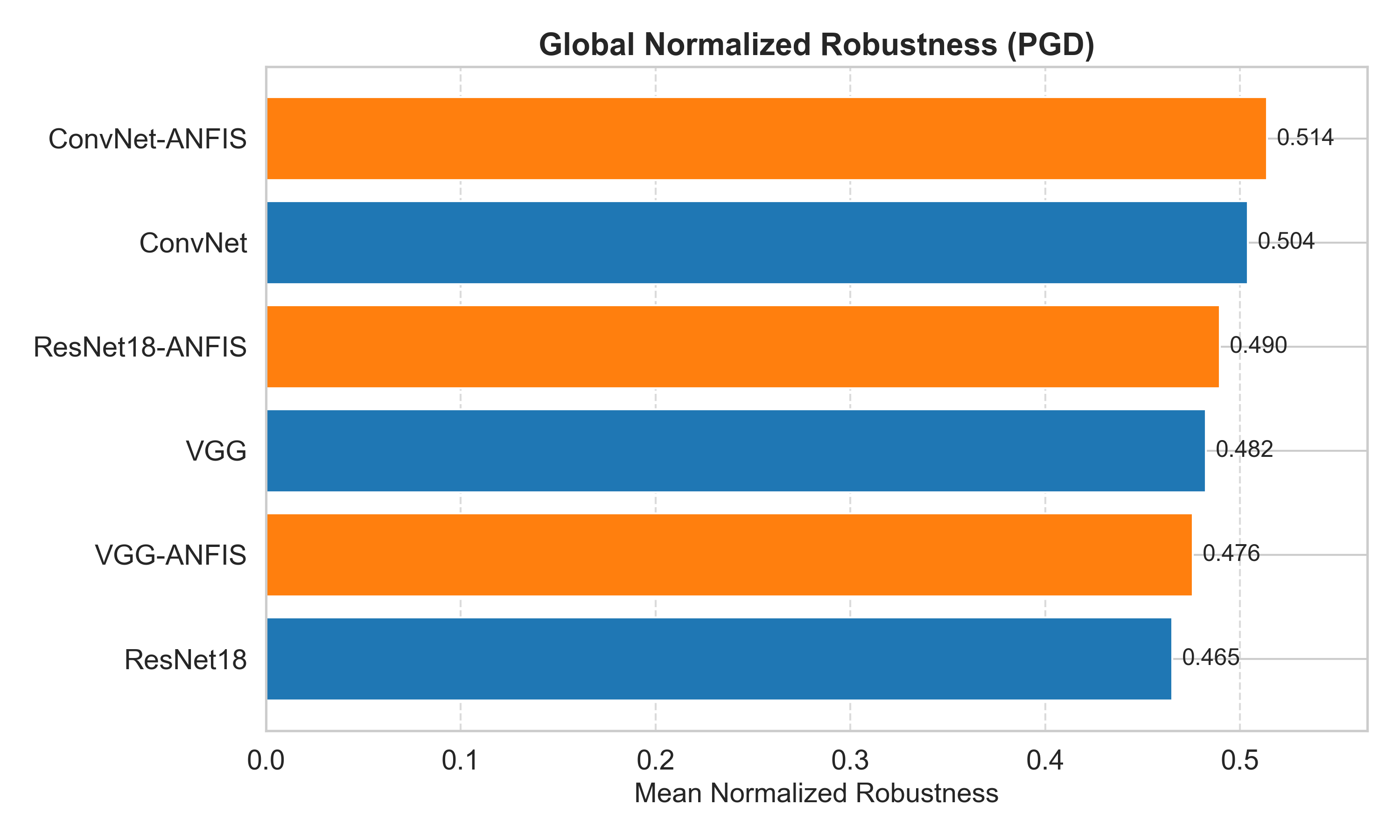}
    \caption{Normalized Robustness (excluding CIFAR-100) for PGD Attacks}
    \label{fig:normalized_robustness_pgd_without_cifar100}
\end{figure}

\begin{figure}[htbp]
    \centering
    \includegraphics[width=0.9\textwidth]{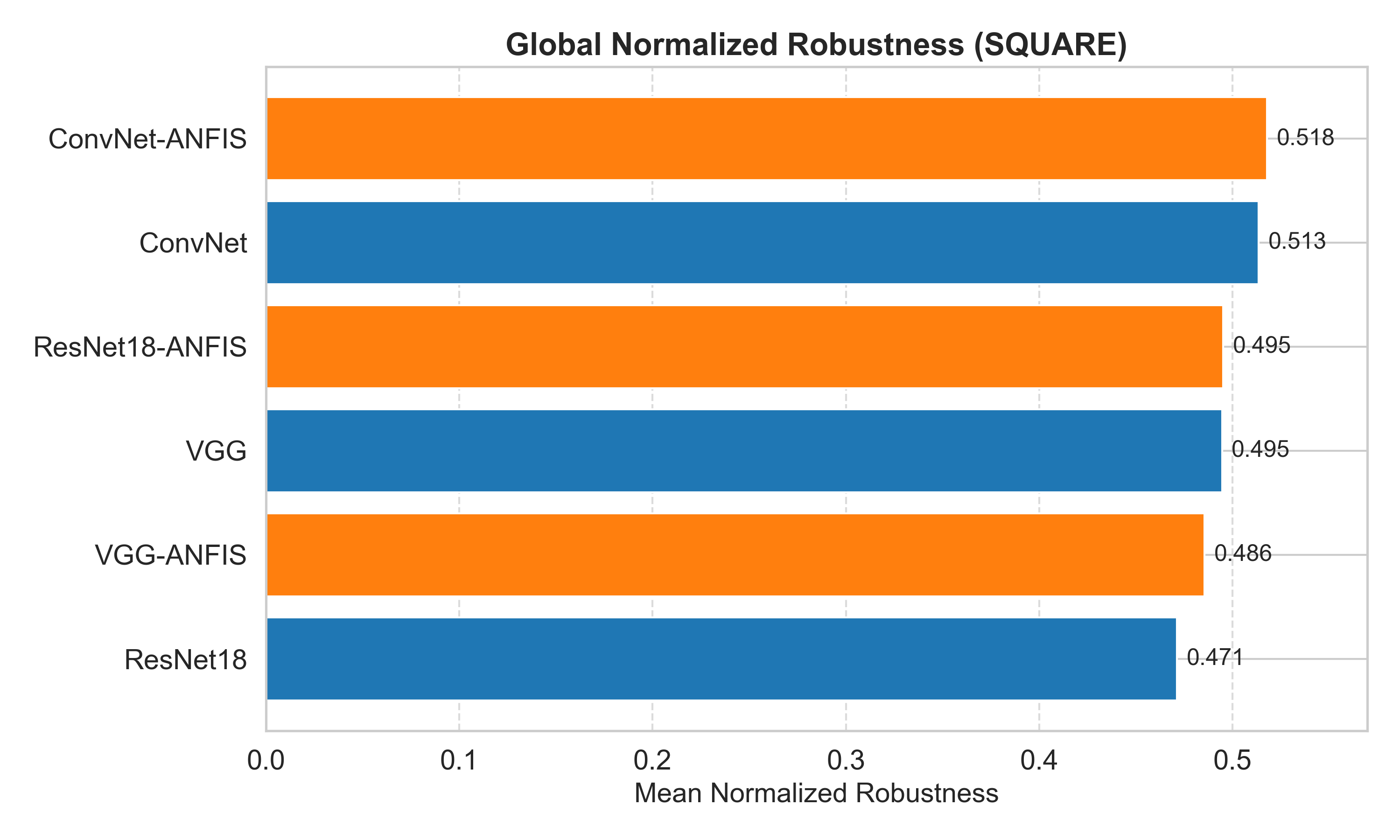}
    \caption{Normalized Robustness (excluding CIFAR-100) for Square Attacks}
    \label{fig:normalized_robustness_square_without_cifar100}
\end{figure}

\FloatBarrier
\section{Discussion}
The results in Table \ref{tab:accuracy_comparison} indicate strong performance across all models on MNIST and Fashion-MNIST, where accuracies are uniformly high and differences between architectures—both with and without ANFIS augmentation—are relatively small. As dataset complexity increases in CIFAR10 and CIFAR100, the performance gaps between models become more pronounced; deeper architectures such as VGG and ResNet18 generally outperform the simpler ConvNet. However, the addition of ANFIS does not yield a consistent improvement across datasets or architectures. While ANFIS-enhanced variants occasionally produce modest gains (for example VGG-ANFIS on CIFAR100), in other cases the hybrid models underperform their baseline counterparts (ConvNet-ANFIS on CIFAR10 and CIFAR100). This mixed pattern suggests that the integration of ANFIS does not systematically enhance visual recognition performance.

Figure \ref{fig:normalized_robustness_pgd} illustrates that the impact of ANFIS integration on PGD robustness is highly dependent on the underlying convolutional architecture. In the case of ConvNet, the ANFIS-enhanced model achieves adversarial accuracy comparable to its baseline counterpart, indicating little measurable benefit. For ResNet18, however, the ANFIS variant demonstrates a noticeable improvement of approximately five percentage points, suggesting that more residual-based architectures may better leverage the additional fuzzy-inference layer. Conversely, VGG displays the opposite trend: the baseline model consistently outperforms VGG-ANFIS under PGD perturbations, suggesting that ANFIS integration may be less effective for architectures lacking strong feature-preservation mechanisms such as those provided by residual connections in ResNet. This same pattern is reinforced in Figure~\ref{fig:normalized_robustness_square}, where we see that ANFIS variants follow architecture-dependent trends: models like ResNet18-ANFIS exhibit clear robustness gains, while VGG-ANFIS continues to underperform its baseline counterpart.

Without CIFAR-100, the robustness trends become much clearer. As shown in Figures~\ref{fig:normalized_robustness_pgd_without_cifar100} and~\ref{fig:normalized_robustness_square_without_cifar100}, the gap between the generic models and their ANFIS variants increases slightly, but the overall pattern remains consistent.

\FloatBarrier
\section{Future Work}
Future work will explore stronger and adaptive adversarial attacks that explicitly target the neuro-fuzzy components of DCNFIS, including transfer-based and expectation-over-transformation attacks. We also plan to investigate robustness-improving strategies such as adversarial training and regularization of fuzzy rules and membership functions. Additionally, studying how design choices in the ANFIS layer, including the number of rules and membership parameterization, affect adversarial robustness remains an important direction. Finally, extending this evaluation to larger datasets and safety-critical applications will help assess the practical robustness of neuro-fuzzy vision models.

\FloatBarrier
\begin{credits}
\subsubsection{\ackname}
The authors extend their sincere gratitude to the members of the AI Bio Lab at the University of Cincinnati for their invaluable discussions and collaborative efforts that facilitated the realization of this work.
\subsubsection{\discintname}
The authors have no competing interests to disclose.\end{credits}

\FloatBarrier
\bibliographystyle{unsrt}
\bibliography{Bib}

@article{MAMDANI19751,
title = {An experiment in linguistic synthesis with a fuzzy logic controller},
journal = {International Journal of Man-Machine Studies},
volume = {7},
number = {1},
pages = {1-13},
year = {1975},
issn = {0020-7373},
doi = {https://doi.org/10.1016/S0020-7373(75)80002-2},
url = {https://www.sciencedirect.com/science/article/pii/S0020737375800022},
author = {E.H. Mamdani and S. Assilian}
}

@ARTICLE{TSK,
  author={Takagi, Tomohiro and Sugeno, Michio},
  journal={IEEE Transactions on Systems, Man, and Cybernetics}, 
  title={Fuzzy identification of systems and its applications to modeling and control}, 
  year={1985},
  volume={SMC-15},
  number={1},
  pages={116-132},
  doi={10.1109/TSMC.1985.6313399}}

@article{ZADEH1965338,
title = {Fuzzy sets},
journal = {Information and Control},
volume = {8},
number = {3},
pages = {338-353},
year = {1965},
issn = {0019-9958},
doi = {https://doi.org/10.1016/S0019-9958(65)90241-X},
url = {https://www.sciencedirect.com/science/article/pii/S001999586590241X},
author = {L.A. Zadeh}}

@article{Castro1995,
  author    = {L. I. Castro},
  title     = {Fuzzy logic controllers are universal approximators},
  journal   = {IEEE Transactions on Systems, Man, and Cybernetics},
  year      = {1995},
  volume    = {25},
  number    = {4},
  pages     = {629--635},
  doi       = {10.1109/21.370193}
}

@article{ANFIS,
author = {Jang, Jyh-Shing},
year = {1993},
month = {06},
pages = {665 - 685},
title = {ANFIS Adaptive-Network-based Fuzzy Inference System},
volume = {23},
journal = {Systems, Man and Cybernetics, IEEE Transactions on},
doi = {10.1109/21.256541}
}

@ARTICLE{LeNet,
  author={Lecun, Y. and Bottou, L. and Bengio, Y. and Haffner, P.},
  journal={Proceedings of the IEEE}, 
  title={Gradient-based learning applied to document recognition}, 
  year={1998},
  volume={86},
  number={11},
  pages={2278-2324},
  doi={10.1109/5.726791}}

@inproceedings{alexnet,
 author = {Krizhevsky, Alex and Sutskever, Ilya and Hinton, Geoffrey E},
 booktitle = {Advances in Neural Information Processing Systems},
 editor = {F. Pereira and C.J. Burges and L. Bottou and K.Q. Weinberger},
 pages = {},
 publisher = {Curran Associates, Inc.},
 title = {ImageNet Classification with Deep Convolutional Neural Networks},
 url = {https://proceedings.neurips.cc/paper_files/paper/2012/file/c399862d3b9d6b76c8436e924a68c45b-Paper.pdf},
 volume = {25},
 year = {2012}
}

@misc{resnet,
  title        = {Deep Residual Learning for Image Recognition},
  author       = {He, Kaiming and Zhang, Xiangyu and Ren, Shaoqing and Sun, Jian},
  year         = {2015},
  eprint       = {1512.03385},
  archivePrefix= {arXiv},
  primaryClass = {cs.CV},
  note         = {Available at \url{https://arxiv.org/abs/1512.03385}},
}

@misc{densenet,
      title={Densely Connected Convolutional Networks}, 
      author={Gao Huang and Zhuang Liu and Laurens van der Maaten and Kilian Q. Weinberger},
      year={2018},
      eprint={1608.06993},
      archivePrefix={arXiv},
      primaryClass={cs.CV},
      url={https://arxiv.org/abs/1608.06993}, 
        note={Available at \url{https://arxiv.org/abs/1608.06993}}
}

@misc{vgg,
      title={Very Deep Convolutional Networks for Large-Scale Image Recognition}, 
      author={Karen Simonyan and Andrew Zisserman},
      year={2015},
      eprint={1409.1556},
      archivePrefix={arXiv},
      primaryClass={cs.CV},
      url={https://arxiv.org/abs/1409.1556}, 
        note={Available at \url{https://arxiv.org/abs/1409.1556}}
}

@misc{dcnfis,
      title={DCNFIS: Deep Convolutional Neuro-Fuzzy Inference System}, 
      author={Mojtaba Yeganejou and Kimia Honari and Ryan Kluzinski and Scott Dick and Michael Lipsett and James Miller},
      year={2024},
      eprint={2308.06378},
      archivePrefix={arXiv},
      primaryClass={cs.AI},
      url={https://arxiv.org/abs/2308.06378}, 
        note={Available at \url{https://arxiv.org/abs/2308.06378}}

}

@misc{adversarialsurvey,
      title={Adversarial Attacks and Defenses in Machine Learning-Powered Networks: A Contemporary Survey}, 
      author={Yulong Wang and Tong Sun and Shenghong Li and Xin Yuan and Wei Ni and Ekram Hossain and H. Vincent Poor},
      year={2023},
      eprint={2303.06302},
      archivePrefix={arXiv},
      primaryClass={cs.LG},
      url={https://arxiv.org/abs/2303.06302}, 
        note={Available at \url{https://arxiv.org/abs/2303.06302}}
}

@misc{pgd,
      title={Towards Deep Learning Models Resistant to Adversarial Attacks}, 
      author={Aleksander Madry and Aleksandar Makelov and Ludwig Schmidt and Dimitris Tsipras and Adrian Vladu},
      year={2019},
      eprint={1706.06083},
      archivePrefix={arXiv},
      primaryClass={stat.ML},
      url={https://arxiv.org/abs/1706.06083}, 
        note={Available at \url{https://arxiv.org/abs/1706.06083}}
}

@misc{squareattack,
      title={Square Attack: a query-efficient black-box adversarial attack via random search}, 
      author={Maksym Andriushchenko and Francesco Croce and Nicolas Flammarion and Matthias Hein},
      year={2020},
      eprint={1912.00049},
      archivePrefix={arXiv},
      primaryClass={cs.LG},
      url={https://arxiv.org/abs/1912.00049}, 
        note={Available at \url{https://arxiv.org/abs/1912.00049}}

}

@article{kim2020torchattacks,
title={Torchattacks: A pytorch repository for adversarial attacks},
author={Kim, Hoki},
journal={arXiv preprint arXiv:2010.01950},
year={2020}
}

@article{fashion,
  author       = {Han Xiao and
                  Kashif Rasul and
                  Roland Vollgraf},
  title        = {Fashion-MNIST: a Novel Image Dataset for Benchmarking Machine Learning
                  Algorithms},
  journal      = {CoRR},
  volume       = {abs/1708.07747},
  year         = {2017},
  url          = {http://arxiv.org/abs/1708.07747},
  eprinttype    = {arXiv},
  eprint       = {1708.07747},
  bibsource    = {dblp computer science bibliography, https://dblp.org}
}

@Techreport{krizhevsky2009learning,
 author = {Krizhevsky, Alex and Hinton, Geoffrey},
 address = {Toronto, Ontario},
 institution = {University of Toronto},
 publisher = {University of Toronto},
 title = {Learning multiple layers of features from tiny images},
 year = {2009},
 title_with_no_special_chars = {Learning multiple layers of features from tiny images},
 url = {https://www.cs.toronto.edu/~kriz/learning-features-2009-TR.pdf}
}

@article{Ujong2025Developmen,
  author = {Jesam Abam Ujong and Fortune K. C. Onyelowe and Ekeoma Emmanuel and M. Vishnupriyan and Kizito C. Ibe and Benjamin Ugorji and Chidobere Nwa-David and Light Ihenna and Jesuborn Obimba-Wogu},
  title = {Development of an Adaptive Neuro-fuzzy Inference System (ANFIS) for Predicting Pavement Deterioration},
  journal = {Sustainable Intelligent Infrastructure},
  year = {2025},
  volume = {1},
  number = {1},
  pages = {39-51},
  doi = {10.62762/SII.2025.494563},
  url = {https://www.icck.org/article/abs/SII.2025.494563},
  issn = {3067-8137},
  publisher = {Institute of Central Computation and Knowledge}
}

@article{yabana2025comparative,
  title={A comparative study of neuro-fuzzy and neural network models in predicting length of stay in university hospital},
  author={Yabana Kiremit, Birg{\"u}l and Dikmeta{\c{s}} Yardan, Elif},
  journal={BMC Health Services Research},
  volume={25},
  number={1},
  pages={446},
  year={2025},
  publisher={Springer}
}

@article{yesmin2025fuzzy,
  title={A Fuzzy Logic-Based Framework for Explainable Machine Learning in Big Data Analytics},
  author={Yesmin, Farjana and Shirmin, Nusrat},
  journal={arXiv preprint arXiv:2510.05120},
  year={2025}
}

@article{brosolo2025road,
  title={The Road Less Traveled: Investigating Robustness and Explainability in CNN Malware Detection},
  author={Brosolo, Matteo and Puthuvath, Vinod and Conti, Mauro},
  journal={arXiv preprint arXiv:2503.01391},
  year={2025}
}

@inproceedings{pavlitska2025robust,
  title={Robust Experts: the Effect of Adversarial Training on CNNs with Sparse Mixture-of-Experts Layers},
  author={Pavlitska, Svetlana and Fan, Haixi and Ditschuneit, Konstantin and Z{\"o}llner, J Marius},
  booktitle={Proceedings of the IEEE/CVF International Conference on Computer Vision},
  pages={251--260},
  year={2025}
}

@article{villegas2024evaluating,
  title={Evaluating the robustness of deep learning models against adversarial attacks: An analysis with fgsm, pgd and cw},
  author={Villegas-Ch, William and Jaramillo-Alc{\'a}zar, Angel and Luj{\'a}n-Mora, Sergio},
  journal={Big Data and Cognitive Computing},
  volume={8},
  number={1},
  pages={8},
  year={2024},
  publisher={MDPI}
}

@inproceedings{waghela2024robust,
  title={Robust image classification: Defensive strategies against FGSM and PGD adversarial attacks},
  author={Waghela, Hetvi and Sen, Jaydip and Rakshit, Sneha},
  booktitle={2024 Asian Conference on Intelligent Technologies (ACOIT)},
  pages={1--7},
  year={2024},
  organization={IEEE}
}

@article{dai2024adversarial,
  title={An adversarial example attack method based on predicted bounding box adaptive deformation in optical remote sensing images},
  author={Dai, Leyu and Wang, Jindong and Yang, Bo and Chen, Fan and Zhang, Hengwei},
  journal={PeerJ Computer Science},
  volume={10},
  pages={e2053},
  year={2024},
  publisher={PeerJ Inc.}
}

\FloatBarrier
\end{document}